\documentclass{article}

\usepackage{times}
\usepackage{graphicx,subfigure}

\usepackage{natbib}

\usepackage{algorithm} 
\usepackage{algorithmic}

\usepackage{hyperref}

\usepackage[accepted]{icml2017}
\usepackage{amsmath, mathtools, amssymb}

\icmltitlerunning{Systematic Testing of Convolutional Neural Networks for Autonomous Driving}

\newcommand{\vm}{\mathbf{m}}

\newcommand{\vx}{\mathbf{x}}
\newcommand{\vy}{\mathbf{y}}

\newcommand{\featsp}{X}
\newcommand{\featsubsp}{\tilde{\featsp}}
\newcommand{\labsp}{Y}
\newcommand{\modsp}{M}

\newcommand{\genf}{\gamma}

\newcounter{myctr}

\begin{document} 

\twocolumn[
\icmltitle{Systematic Testing of Convolutional Neural Networks for Autonomous Driving}



\icmlsetsymbol{equal}{*}

\begin{icmlauthorlist}
\icmlauthor{Tommaso Dreossi}{ucb}
\icmlauthor{Shromona Ghosh}{ucb}
\icmlauthor{Alberto Sangiovanni-Vincentelli}{ucb}
\icmlauthor{Sanjit A. Seshia}{ucb}
\end{icmlauthorlist}

\icmlaffiliation{ucb}{University of California, Berkeley, USA}

\icmlcorrespondingauthor{Tommaso Dreossi}{dreossi@berkeley.edu}
\icmlcorrespondingauthor{Shromona Ghosh}{shromona.ghosh@berkeley.edu}

\icmlkeywords{Convolutional Neural Network, Testing, Analysis}

\vskip 0.3in
]



\printAffiliationsAndNotice{\icmlEqualContribution} 

\begin{abstract} 
We present a framework to systematically analyze convolutional neural  
networks (CNNs) used in classification of cars in autonomous vehicles. Our analysis procedure comprises an image generator that produces synthetic pictures by sampling in a lower dimension image modification subspace and a suite of visualization 
tools. The image generator produces images which can be used to test the CNN and hence expose its vulnerabilities. The presented framework can be used to extract insights of the CNN classifier, compare across classification models, or generate training and validation datasets.  

\end{abstract} 

\section{Introduction}\label{sec:introduction}

Convolutional neural networks (CNN) are powerful models that
have recently achieved the state of the art in object classification and detection tasks. It is no surprise that they are used extensively in large scale Cyber-Physical Systems (CPS). For CPS used in safety critical purposes, verifying CNN models is of utmost importance~\cite{dreossi2017compositional}.
An emerging domain where CNNs have found application is autonomous driving where
object detectors are used to identify cars, pedestrians, or road signs~\cite{dougherty1995review, BojarskiTDFFGJM16}.

CNNs are usually composed of extensively parallel nonlinear layers that 
allow the networks to learn highly nonlinear functions. While CNNs
are able to achieve high accuracy for object detection, their analysis has proved to be
extremely difficult. Proving their correctness, i.e.,
to show that a CNN always correctly detects a particular object, has become practically impossible. 
One approach to address this problem analyzes the robustness of CNNs with respect to perturbations.
Using optimization-based techniques~\cite{szegedy2013intriguing, papernot2016limitations} or generative NNs~\cite{goodfellow2014generative},
it is possible to find minimal adversarial modifications that can cause a CNN to misclassify an altered picture.
Another approach inspired by formal methods aims at formally proving the correctness of neural networks by using, e.g., Linear Programming or SMT solvers~\cite{huang2016safety, katz2017reluplex}.
Unfortunately, these verification techniques usually impose restrictions on treated CNNs and suffer from scalability issues.

In this work, we present a framework to
systematically test CNNs by generating synthetic datasets.
In contrast to the
adversarial generation techniques, we aim at generating realistic 
pictures rather than introducing perturbations into preexisting ones. In this paper,
we focus on self-driving applications, precisely on CNNs used for detection
of cars. However, the presented techniques are general enough for application to other domains.

Our framework consists of three main modules: an \emph{image generator},
a collection of \emph{sampling methods}, and a suite of \emph{visualization tools}.
The image generator renders realistic images of road scenarios.
The images are obtained by arranging basic objects
(e.g., road backgrounds, cars) and by tuning
image parameters (e.g., brightness, contrast, saturation). By preserving the aspect ratios of the objects, we generate more realistic images. 
All possible configurations of the objects and image parameters define a \emph{modification space} whose elements map to a subset of the CNN feature space (in our case, road scenarios).
The goal of the sampling methods is to provide modification points
to the image generator that produces pictures used to 
extract information from the CNN. We provide different sampling techniques, depending on the 
user needs. In particular, we focus on samplings methods
that cover the modification space evenly and active optimization-based methods to generate images that are misclassified by the analyzed CNN.
Finally, the visualization tools are used to display the gathered information.
Our tool can display the sampled modifications against metrics of interest such as 
the probability associated with the predicted bounding boxes (the box containing the car) or the intersection over union
(IOU) used to measure the accuracy of the prediction box.

The contributions provided by our framework are twofold:
\begin{itemize}
	\item \emph{Analysis of Neural Network Classifiers}. The systematic analysis is
		useful to obtain insights of the considered CNN classifier,
		such as the identification of blind spots or corner cases. 
		Our targeted testing can also be used to
		compare across CNN models;
	\item \emph{Dataset Generator}. Our picture generator can
		generate large data sets for which the diversity
		of the pictures can be controlled by the user.
		This overcomes a lack of training data,
		one of the limiting problems in training of CNNs.
		Also, a target synthesized dataset can be used as a 
		benchmark for a specific domain of application.
\end{itemize}

We present a systematic methodology for finding failure cases of CNN classifiers. This is a first attempt towards verifying machine learnt components in complex systems.

The paper is organized as follows: Sec.~\ref{sec:analyzer} describes the
analysis framework and defines the picture generator, sampling methods,
and visualization tools; Sec.~\ref{sec:evaluation} to implementation
details and experimental evaluations.

\section{CNN Analyzer}\label{sec:analyzer}

\subsection{Overview}

We begin by introducing some basic notions and by giving
an overview of our CNN analysis framework.

Let $f : \featsp \to \labsp$ be a CNN that assigns to every 
\emph{feature vector} $\vx \in \featsp$ a \emph{label} $\vy \in \labsp$,
where $\featsp$ and $\labsp$ are a feature and a label space, respectively.
In our case $\vx$ can be a picture of a car and $\vy$ is the CNN prediction
representing information such as the detected object class, the prediction confidence,
or the object position.

Our analysis technique (Alg.~\ref{algo:analysis}) consists in a 
loop where at each step an image modification configuration $\vm$
is sampled, an image $\vx$ is generated using the modification $\vm$, and
a prediction $\vy$ is returned by the analyzed CNN.
Intuitively, $\vm$ describes the configuration
of the synthetic picture $\vx$ to be generated. $\vy$ is then the prediction of the CNN on this generated image $\vx$. A modification $\vm$
can specify, for instance, the x and y coordinates of a car in a picture as 
well as the brightness or contrast of the image to be generated. 
At each loop iteration, the information $\vm, \vx, \vy$ are stored
in the data structure $D$ that is later used to inspect and visualize the CNN behavior.
The loop is repeated until a condition on $D$ is met. Some examples of halting conditions
can be the discovery of a misclassified picture, a maximum number of generated images,
or the achievement of coverage threshold on the modification space.

\begin{algorithm}
	\caption{Analyze CNN}
	\begin{algorithmic}\label{algo:analysis}
	\FUNCTION{CNNanalysis}
	\REPEAT
		\STATE $\vm \gets$ sample($\modsp$)
		\STATE $\vx \gets$ generateImage($\vm$)
		\STATE $\vy \gets f(\vx)$
		\STATE $D.$add($\vm,\vx,\vy$)
	\UNTIL{condition($D$)}
	\STATE visualize($D$)
	\ENDFUNCTION
\end{algorithmic}
\end{algorithm}


The key steps of this algorithm are the picture generation (i.e., how an image is rendered from
a modification choice) and how modifications are sampled (i.e., how to chose a modification 
in such a way to achieve the analysis goal).
In the following, we define image modifications and show how synthetic pictures
are generated (Sec.~\ref{sec:abstraction}). Next, we introduce some sampling methods (Sec.~\ref{sec:sampling}),
and finally some visualization tools (Sec~\ref{sec:visualization}).

\subsection{Image Generation}\label{sec:abstraction}


Let $\featsubsp \subseteq \featsp$ be a subset of the feature
space of $f : \featsp \to \labsp$.
A \emph{generation function}
$\genf : \modsp \to \featsubsp$ is a function that maps every modification  
$\vm \in \modsp$ to a feature $\genf(\vm) \in \featsubsp$.

Modification functions can be used to compactly
represent a subset of the feature space. For instance, modifications
of a given picture, such as displacement of a car and brightness,
can be seen as the dimensions of a $2$-D modification space.
Low-dimensional modification spaces allow us to analyze CNNs
on compact domains as opposite to intractable feature spaces. 
Let us clarify these concepts with the following example where
a set of pictures is abstracted into a $3$-D box.

Let $\featsp$ be the set of $1242 \times 375$ RBG pictures
(Kitti image resolution~\cite{Geiger2013IJRR}).
Since we are interested in the automotive context, we consider
the subset $\featsubsp \subset \featsp$ of pictures of cars
in different positions on a particular background with different
color contrasts. In these settings, we can define, for instance, 
the generation function $\genf$ that maps
$\modsp = [0,1]^3$ to $\featsubsp$, where the dimensions of
$\modsp$ characterize the $x,y$ positions of the car and
the image contrast, respectively. For instance, 
$\genf(0,0,0)$ places the car on the left close to the observer
with high contrast, $\genf(1,0,0)$ shifts the car to the right,
or $\genf(1,1,1)$ sees the car on the right, far from the observer, with
low contrast.

\begin{figure}
	\centering
	\includegraphics[scale=0.2]{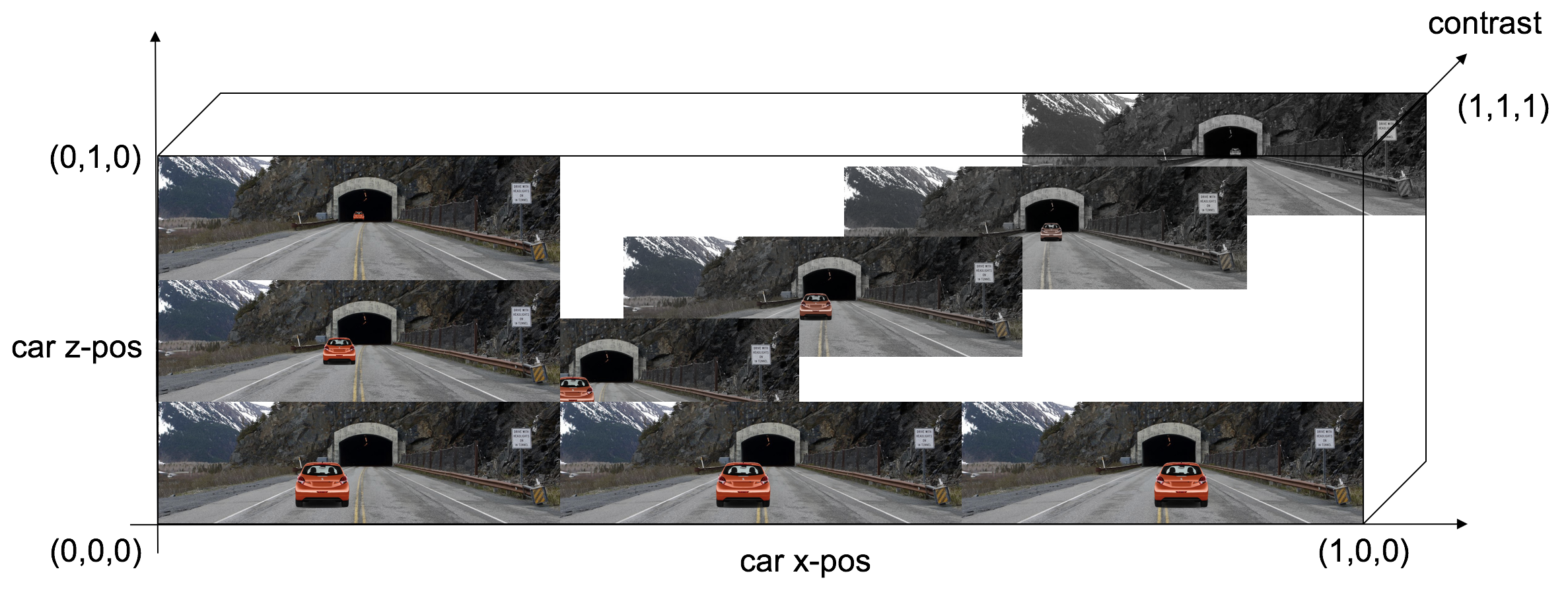}
	\caption{Modification space (surrounding box) and corresponding synthetic images.\label{fig:abssp}}
\end{figure}

Fig.~\ref{fig:abssp} shows some images of $\featsubsp$
disposed accordingly to
their location in the modification space $\modsp$. When moving on the 
$x$-axis of $\modsp$, the car shifts horizontally; a change on the $y$-axis 
affects the position of the car in depth;
the $z$-axis affects the contrast of the picture.
This simple example shows how the considered feature space
can be abstracted in a $3$-D modification space in which every point
corresponds to a particular image.

\begin{figure}
	\centering
	\includegraphics[scale=0.2]{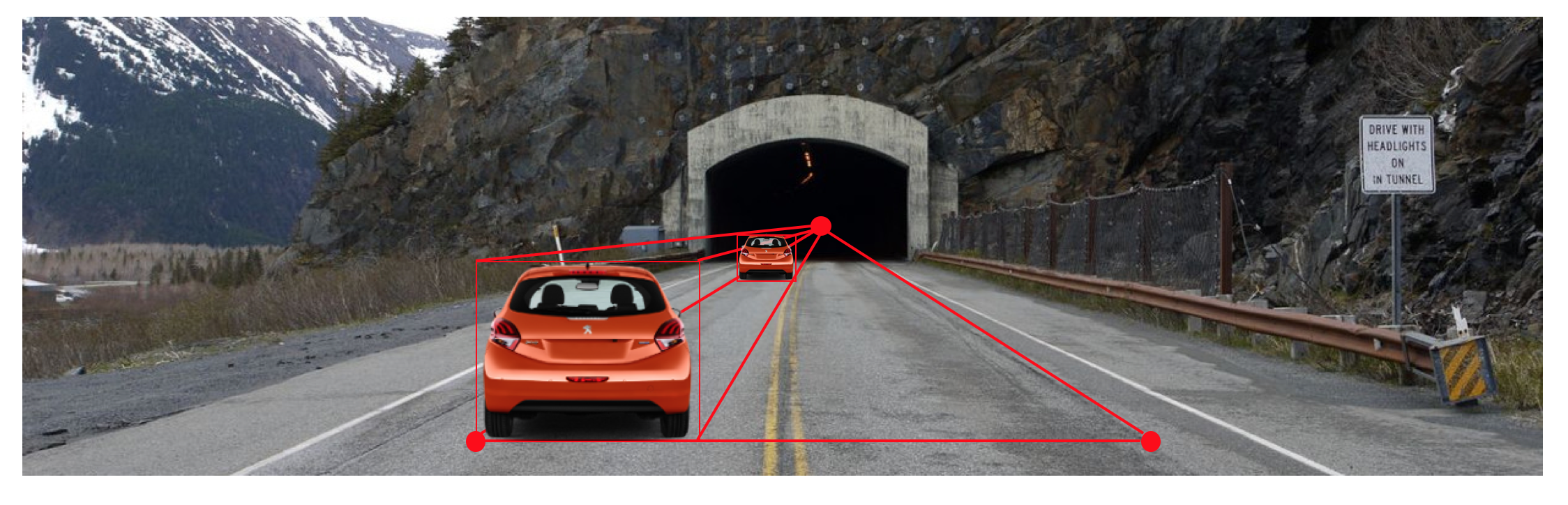}
	\caption{Car resizing and displacement
	using vanishing point and lines.\label{fig:vanish}}
\end{figure}

In this examples, we chose the extreme positions 
of the car (i.e., maximum and minimum $x$ and
$y$ position of the car) on the sidelines of the road and
the image vanishing point. Both the sidelines and the 
vanishing point can be automatically detected~\cite{aly2008real,kong2009vanishing}.
The vanishing point is useful to determine the vanishing lines 
necessary to resize and place the car when 
altering its position in the y modification dimension.
For instance, the car is placed and shrunk towards the vanishing point
as the y coordinate of its modification element gets close to $1$ (see Fig.~\ref{fig:vanish}).

\subsection{Sampling Methods}\label{sec:sampling}

We now consider some methods to efficiently sample the
modification space. Good sampling techniques should 
provide a high coverage of the abstract space and identify
samples whose concretizations lead to misclassifying images.

\paragraph{Low-discrepancy Sequences}

A \emph{low-discrepancy} (or quasi-random) sequence is a sequence of $n$-tuples
that fills an $n$-D space more uniformly than uncorrelated random points.
Low-discrepancy sequences are useful to cover boxes by reducing gaps and clustering of points.

Let $U = [0,1]^n$ be a $n$-D box, $J \subseteq U$ be a sub-box, and $X \subset U$ be a 
set of $m$ points. The \emph{discrepancy} $D(J,X)$ of $J$ is the difference between the
proportion of points in $J$ compared to $U$ and the volume of $J$ compared to $U$:
\begin{equation}
	D(J,X) = |\#(J)/m - vol(J)|
\end{equation}
where $\#(J)$ is the number of points of $X$ in $J$ and $vol(J)$ is the 
volume of $J$.
The \emph{star-discrepancy} $D^*(X)$ is the worst case distribution of $X$:
\begin{equation}
	D^*(X) = \max_J D(J,X)
\end{equation}
Low-discrepancy sequences generate sets of points that minimize the 
star-discrepancy. Some examples of low-discrepancy sequences are
the Van der Corput, Halton~\cite{halton1960efficiency}, or Sobol~\cite{sobol1976uniformly} sequences. 
In our experiments, we used the Halton and lattice-based~\cite{niederreiter1988low} sequences. 
These sampling methods ensure an optimal coverage of the 
abstract space and allows us to identify clusters of misclassified pictures as well as
isolated corner cases otherwise difficult to spot with a uniform sampling technique.

\paragraph{Active Learning}
At every step, given a sample, we generate images which are presented as input to the neural network under test. 
This becomes an expensive process when the number of samples necessary for the covering the input space is large. 
We propose using active learning to minimize the number of images generated and only sample points which have a 
high probability of being a counter example. 

We model the function from the sample space $U = [0,1]^n$ to the score (output) of the CNN as a Gaussian Process (GP).   
GPs are a popular choice for nonparametric regression in machine learning, where the goal is to find an approximation of a nonlinear 
map $p(u): U \rightarrow R$ from an input sample $ u \in  U$ to the score produced by the neural network. The main assumption is that 
the values of $p$, associated with the different sample inputs, are random variables and have a joint Gaussian distribution. 
This distribution is specified by a mean function, which is assumed to be zero without loss of generality, and a covariance function $k(u, u′)$,
called kernel function. 

The GP framework can be used to predict the score $p(u)$
at an arbitrary sample $u \in U$ based on a set of $t$ past observations 
$y_t =  [\tilde{p}(u_1), \dots, \tilde{p}(u_t)]^T$ at samples $U_t = \{u_1 , \dots , u_t\}$ without generating the image for $u$. The observations 
of the function values $\tilde{p}(u_t ) = p(u_t ) + w_t$ are corrupted by Gaussian noise $w_t \sim N (0, \sigma^2)$. Conditioned on these 
observations, the mean and variance of the prediction at $u$ are given by: 
\begin{equation}
\begin{split}
	\mu_t(u)  &= k_t(u)(K_t+ \mathbb{I}_t\sigma^2)^{-1} y_t \\
	\sigma_t^1(u) &=k(u,u)- k_t(u)(K_t+ \mathbb{I}_t \sigma^2)^{-1} k_t^T(u) \\
\end{split}
\end{equation}
where the vector $k_t(u)= [k(u,u_1),\dots,k(u,u_t)]$ contains the covariances between the new input, $u   $, and the past data points in $U_t$, 
the covariance matrix, $K_t \in  \mathbb{R}^{t\times t}$, has entries $[K_t](i,j) = k(u_i,u_j)$ for $i,j \in \{1,\dots,t\}$, and the identity matrix 
is denoted by $\textbf{I}_t  \in \mathbb{R}^{t\times t}$.
	
	Given a GP, any Bayesian optimization algorithm is designed to find the global optimum of an unknown function within few evaluations on 
	the real system. Since we search for counterexamples, i.e, samples where the score returned by the neural network is low, we use \textit{GP-Lower Confidence Bound} (GP-LCB) as our objective function. Since the optimal sample $u_t^*$ is not known a priori, the optimal strategy has to balance learning about the location of the most falsifying sample (exploration), and selecting a sample that is known to lead to low scores (exploitation). We formulate the objective function as, 
		$u_t = \text{argmin}_{u \in U} \mu_{t-1}(u) - \beta^{1/2}_t\sigma_{t-1}(u)$
	where $\beta_t$ is a constant which determines the confidence bound.


\subsection{Visualization}\label{sec:visualization}

We now show how the gathered information can be visualized and interpreted.

\begin{figure}
\centering     
\subfigure[SqueezeDet.]{\label{fig:squeeze_map}\includegraphics[scale=0.15]{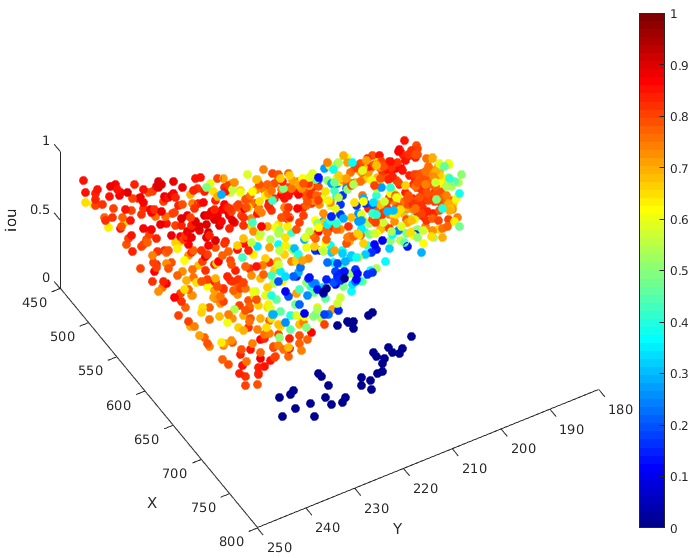}}
\subfigure[Yolo.]{\label{fig:squeeze_map}\includegraphics[scale=0.15]{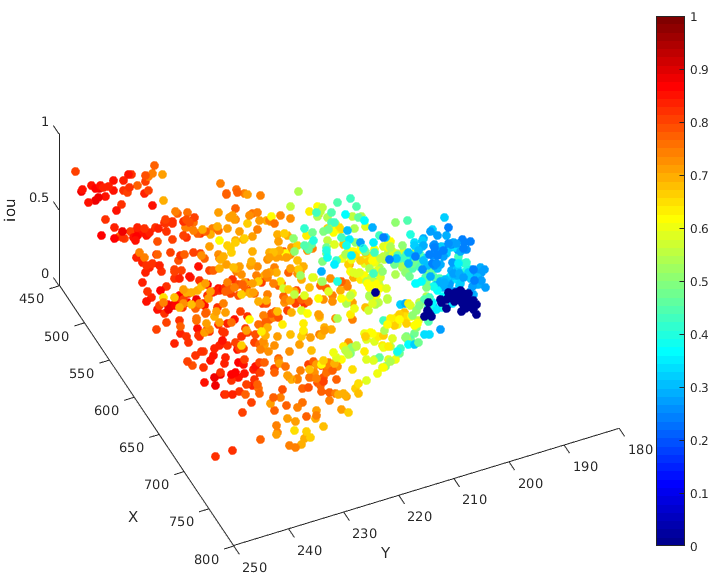}}
\caption{CNN analysis showing car coordinates ($x,y$), IOU ($z$), and confidence (color).\label{fig:heat_maps}}
\end{figure}

\begin{figure}
\centering     
\subfigure[SqueezeDet.]{\label{fig:squeeze_super}\includegraphics[scale=0.17]{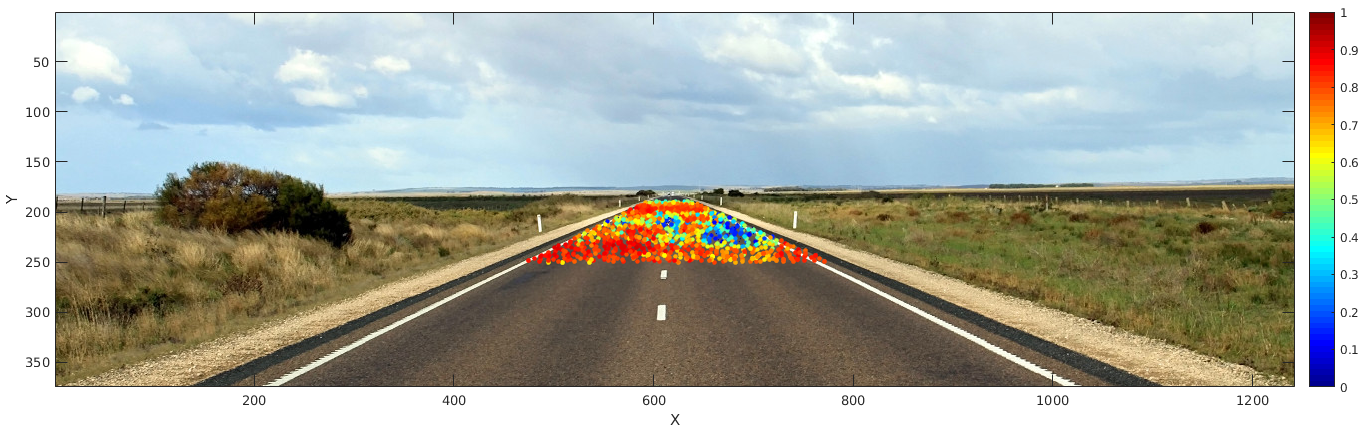}}
\subfigure[Yolo.]{\label{fig:yolo_super} \includegraphics[scale=0.17]{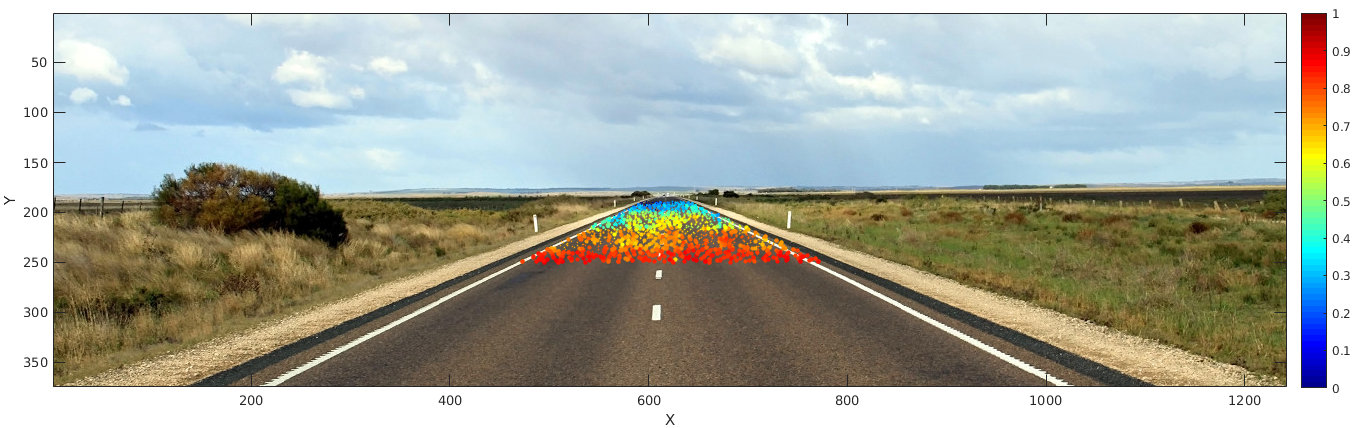}}
\caption{Superimposition of the analysis on road background.\label{fig:superimpose}}
\end{figure}

In our data analysis, we consider two factors: the \emph{confidence score} and the \emph{Intersection Over Union} (IOU) that is
a metric used to measure the accuracy of detections. IOU is defined as the area of 
overlap over the area of the union of the predicted and ground-truth bounding boxes.
Our visualization tool associates the center of the car of the generated images
to the confidence score and IOU returned by the treated CNN. Fig.~\ref{fig:heat_maps} depicts some examples where
the $x$ and $y$ are the center coordinates of the car,
$z$ is the IOU, and the color represents
the CNN confidence score. We also offer the possibility to superimpose the experimental data on the background 
used to render the pictures (see Fig.~\ref{fig:superimpose}). In this case, the IOU is represented by the dimension of the marker.
This representation helps us to identify particular regions of interest on the road. In the next section, we will see how these
data can be interpreted.

\section{Implementation and Evaluation}\label{sec:evaluation}

We implemented the presented framework in a tool available at
\url{https://github.com/shromonag/FalsifyNN}. The tool comes 
with a library composed by a dozen of road backgrounds and car models,
and it interfaces to the CNNs SqueezeDet~\cite{squeezedet}, KittiBox~\cite{TeichmannWZCU16}, and Yolo~\cite{redmon2016you}.
Both the image library and CNN interfaces can be personalized by the user.

As an illustrative case study, we considered a countryside background and a Honda Civic and we 
generated 1k synthetic images using our rendering techniques (Sec.~\ref{sec:abstraction}) and the Halton
sampling sequence (Sec.~\ref{sec:sampling}). We used the generated pictures to analyze 
SqueezeDet~\cite{squeezedet}, a CNN for object detection for autonomous driving, and Yolo~\cite{redmon2016you}, 
a multipurpose CNN for real-time detection.

Fig.~\ref{fig:heat_maps} displays the center of the car in the generated pictures associated with the 
confidence score and IOU returned by both SqueezeDet and Yolo.
Fig.~\ref{fig:superimpose} superimposes the heat maps of Fig.~\ref{fig:heat_maps} on the used background.

There are several interesting insights that emerge from the graphs obtained(for this combination of background and car model).
SqueezeDet has, in general, a high confidence and IOU, but has a blind spot for cars 
in the middle of the road on the right(see the cluster of blue points in Fig.~\ref{fig:squeeze_map} and~\ref{fig:squeeze_super}).
Yolo's confidence and IOU decrease with the car distance (see Fig.~\ref{fig:squeeze_map}). We were able to detect a blind area in Yolo, corresponding to cars on the far left (see blue points in Fig.~\ref{fig:squeeze_map}).


Note how our analysis can be used to visually compare the two CNNs
by graphically highlighting their differences in terms of detections, confidence scores, and IOUs.
A comprehensive analysis and comparison of these CNNs
should involve images generated by combinations of different cars and backgrounds. However, this experiment already shows the benefits 
in using the presented framework and highlights the quantity and quality of information that can be extracted from a 
CNN even with a simple study.

\section*{Acknowledgement}
The authors acknowledge Forrest Iandola and Kurt Keutzer
for giving the presentation of this work at Reliable Machine Learning in the Wild - ICML 2017 Workshop.

\nocite{langley00}

\bibliography{biblio_tom}
\bibliographystyle{icml2017}

\end{document}